\documentclass{article}
\usepackage{ijcai25}

\usepackage{times}
\usepackage{soul}
\usepackage{url}
\usepackage[hidelinks]{hyperref}
\usepackage[utf8]{inputenc}
\usepackage[small]{caption}
\usepackage{graphicx}
\usepackage{amsmath}
\usepackage{amsthm}
\usepackage{booktabs}
\usepackage{algorithm}
\usepackage{algorithmic}
\usepackage[switch]{lineno}
\usepackage{amsfonts}

\usepackage[
singlelinecheck=false 
]{caption}

\usepackage{tabularx}
\newcolumntype{Y}{>{\centering\arraybackslash}X}

\usepackage{subfigure}
\usepackage{graphicx, caption, subcaption}
\usepackage[toc,page]{appendix}
\usepackage{subcaption}
\usepackage{array}
\usepackage{multirow}
\usepackage{booktabs}
\usepackage{makecell}

\title{Signals from the Floods: AI-Driven Disaster Analysis through Multi-Source Data Fusion}

\author{
Xian Gong$^1$
\and
Paul X.~McCarthy$^{1,2}$\and
Lin Tian$^1$\And
Marian-Andrei Rizoiu$^1$\\
\affiliations
$^1$University of Technology Sydney\\
$^2$University of New South Wales\\
\emails
xian.gong@student.uts.edu.au,
paul@onlinegravity.com
\{lin.tian-3, Marian-Andrei.Rizoiu\}@uts.edu.au
}

\begin{document}

\maketitle

\begin{abstract}
Massive and diverse web data are increasingly vital for government disaster response, as demonstrated by the 2022 floods in New South Wales (NSW), Australia. This study examines how X (formerly Twitter) and public inquiry submissions provide insights into public behaviour during crises. We analyse more than 55,000 flood-related tweets and 1,450 submissions to identify behavioural patterns during extreme weather events. While social media posts are short and fragmented, inquiry submissions are detailed, multi-page documents offering structured insights. Our methodology integrates Latent Dirichlet Allocation (LDA) for topic modelling with Large Language Models (LLMs) to enhance semantic understanding. LDA reveals distinct opinions and geographic patterns, while LLMs improve filtering by identifying flood-relevant tweets using public submissions as a reference. This Relevance Index method reduces noise and prioritizes actionable content, improving situational awareness for emergency responders. By combining these complementary data streams, our approach introduces a novel AI-driven method to refine crisis-related social media content, improve real-time disaster response, and inform long-term resilience planning.

\end{abstract}

\section{Introduction}
This study presents new research that extends analytical work originally commissioned by the NSW Government to examine the 2022 floods. It builds on those insights by integrating machine learning techniques to improve real-time disaster response and long-term resilience planning. While the independent NSW Flood Inquiry provided key recommendations, our work aligns with both its findings and the government’s subsequent response, contributing to efforts to enhance disaster communication, emergency coordination, and flood risk management. By leveraging machine learning for tweet filtering, this study ensures that relevant disaster-related content is prioritized, further demonstrating the potential of AI-driven methods in crisis management and policy planning.

Floods account for 43\% of weather disasters (1995–2015), affecting 2.3 billion people \cite{UNISDR_CRED_2015}. Between February and April 2022, catastrophic floods devastated Northern New South Wales (NSW) and parts of South-East Queensland, impacting over 70 local government areas. The disaster tragically claimed 22 lives and caused over \$5 billion in insured losses, making it the costliest flood in Australian history \cite{InsuranceCouncilofAustralia2022}. Sydney, Australia’s most populous city, was also significantly affected by related storms and flooding during this period. In the most affected areas, such as Lismore and Richmond Valley, the devastating flood disaster compelled thousands of people to seek emergency accommodation, damaged thousands of homes, and disrupted business activities, underscoring its profound social, financial, and economic impacts \cite{LismoreCityCouncil2022,RichmondValleyCouncil2022,QueenslandGovernment2022,FloodInquiry2022}. The NSW Premier commissioned the independently-led NSW Floods Inquiry to investigate the causes, assess emergency responses, and recommend future flood management strategies. The inquiry conducted extensive consultations—holding 144 meetings and receiving 1,494 submissions—to inform recommendations on future flood risk management and coordination among government agencies \cite{FloodInquiry2022}. With climate change intensifying natural disasters, understanding public responses, crisis communication, and societal impact is increasingly critical.

Previous studies on disaster response and resilience face two major limitations. First, they often rely solely on social media posts—mostly Twitter—which are inherently short and fragmented, limiting the breadth of analysis. Second, their text analysis typically uses lexicon-based methods that focus mainly on sentiment while neglecting topic extraction. As a result, the full range of insights available from these posts is not fully exploited. This paper introduces a novel analytics framework to analyse the flood disaster by integrating social media and public submissions. Social media complement data from public submissions to flood inquiries by providing real-time data that captures immediate reactions, concerns, and trends during the event. Multiple data sources analysis enables us to explore public behaviour patterns through different lenses, making the results more comprehensive and robust. We separately analyse the differences in public topics reflected in tweets and public submissions and leverage semantic and syntactic information from large language models (LLMs) and public submissions to quantify the relevance of tweets to flood disasters. We aim to answer the following research questions:
\begin{enumerate}
    \item How do topics discussed in social media compare to those in public submissions regarding floods?
    \item How can machine learning (ML) and large language models (LLMs) improve filtering and relevance assessment of flood-related tweets?
\end{enumerate}

Our findings highlight that each medium provides unique perspectives due to its inherent nature and interacts with each other to provide a more comprehensive analysis. \textbf{RQ1.} There are distinct differences between social media discussions and public submissions regarding floods. Public submissions, primarily from flood-affected residents, emergency personnel, and organizations, focus on personal experiences, infrastructure failures, emergency response effectiveness, and long-term policy recommendations. In contrast, social media discussions are more immediate, diverse, and politically charged, covering real-time weather updates, rescue efforts, donation appeals, emotional reactions, and government criticism. While submissions provide structured, in-depth reflections, tweets offer rapid, high-volume insights but contain significant noise and off-topic content. \textbf{RQ2.} To enhance relevance filtering of flood-related tweets, we developed a machine learning-based Relevance Index using LongT5 embeddings. By leveraging public submissions as a reference corpus, we trained an LLM to score tweets based on their semantic similarity to verified flood-related content. This approach effectively filters out unrelated tweets, distinguishing meaningful crisis information from general online chatter. Our results demonstrate that ML and LLMs significantly improve disaster-related information by reducing misinformation, prioritizing critical content, and enhancing real-time situational awareness. These findings underscore the importance of integrating structured public submissions with unstructured social media data to create a more reliable and actionable framework for flood disaster response and policy planning.

This paper makes several key contributions to flood disaster response and crisis management. It introduces a multi-source framework to analyse public behavior during floods. Using Latent Dirichlet Allocation (LDA), it compares real-time, diverse social media discussions with structured, policy-focused public submissions. A machine learning-based Relevance Index is developed to filter flood-related tweets, leveraging public submissions as a reference corpus to improve the signal-to-noise ratio in social media data. By fusing traditional and digital data sources, this study enhances disaster preparedness, real-time crisis response, and long-term resilience planning, providing actionable insights for policymakers, emergency responders, and urban planners.

\section{Related Work}

Literature on flood disaster response and management spans emergency response management, resource allocation and decision-making strategies during floods, and community resilience and recovery, which address rebuilding efforts and psychosocial support \cite{burstein2008decision,cutter2008place}.  Especially related to our work, the role of social media and crowdsourced information in providing real-time situational awareness and enhancing public communication has also been a significant area of investigation \cite{vieweg2010microblogging,fohringer2015social,kryvasheyeu2015performance}. 

The tools and technologies used in flood disaster response and management depend largely on the nature of the data sources. Therefore, we focus on discussing the data sources, social media, public submissions, and corresponding analytical methods involved in our research.

\subsection{Social Media Analysis}

Social networks, particularly X (Twitter), has been recognised as a valuable tool for real-time disaster response in spatiotemporal and sentiment analysis. During Hurricane Sandy, Kryvasheyeu et al. \cite{kryvasheyeu2016rapid} found that social media data could help rapidly assess spatial distribution damage. Subsequent research by Kankanamge et al. \cite{kankanamge2020determining} used geo-located messages from Twitter to demarcate highly impacted disaster zones. The ability to map affected areas by crowdsourcing information from residents complements conventional remote sensing and field assessments, especially in places where official monitoring networks are sparse. 

Studies such as de Albuquerque et al. \cite{de2015geographic} used social media content to assess how information demand and supply evolve over time. They noted that different stages of a disaster require different types of information—such as warnings during the onset of the event, resource needs during peak flooding, and recovery support after the floodwaters recede. As expressed by social media users, this dynamic information flow gives authorities insights into changing community needs, which is vital for effective response.

Sentiment analysis is another promising area in social media analysis for flood disaster management. Studies such as those by Yu et al. \cite{yang2019twitter} employed sentiment analysis to gauge the emotional responses of affected individuals during different stages of flood events. Understanding emotional dynamics can aid disaster management agencies in identifying particularly distressed communities that may require additional support. For example, Twitter sentiment during a flood can highlight hotspots of anxiety or panic, which could indicate an urgent need for intervention. 

While social media analysis offers real-time data, researchers have highlighted the challenges of data quality and misinformation. Imran et al. \cite{imran2015processing} explored extracting ``information nuggets" from social media to enhance response strategies. Their study emphasised that the automated filtering of useful content from large volumes of social media posts can significantly improve the efficiency of emergency response efforts. Plotnick et al. \cite{plotnick2016barriers} discussed the issue of credibility in social media data, emphasising that not all content shared during a flood event can be trusted. Verification of information remains a critical task, with methods such as cross-referencing social media posts with authoritative data sources being recommended to filter out misleading information. 

\subsection{Public Submissions Analysis}

Public submissions (surveys) provide qualitative insights often unattainable through traditional data sources or social media, particularly regarding the socio-economic impacts of floods on affected communities. Studies \cite{thieken2007coping,bubeck2018helps} used surveys to assess economic damages in Germany, highlighting the importance of capturing household-level vulnerabilities and property loss. Analytical methods like descriptive statistics and regression analysis are commonly used to examine factors influencing flood preparedness, as shown by Bubeck et al. \cite{bubeck2012review}, who studied household behaviour in relation to flood risk awareness and mitigation actions. Structural Equation Modelling (SEM), a more advanced analytical method, has been increasingly adopted to understand the relationships between multiple factors influencing flood resilience and preparedness. Santoro et al. \cite{santoro2023community} applied SEM to identify direct and indirect effects of social, psychological, and institutional factors, such as social cohesion, past flood experiences, and trust in authorities. 

Public submissions and surveys are typically analysed manually. Despite the capability of machine learning (ML) and LLMs to efficiently derive insights from extensive qualitative data, their application in public survey analysis is limited, with scant literature available. Broader adoption of ML could facilitate faster assessments and support improved disaster management.

\section{Methodology}

\subsection{Datasets} \label{datasets}
\textbf{Social media data from X (Twitter).} Flood-related tweets from January to May 2022 are collected using relevant keywords and hashtags, covering peak flood activity and recovery discussions. The keywords and hashtags include Floods NSW, Flooding NSW, Lismore Floods, Sydney Floods, Sydney Storm, ``rain bomb", ``NSW Floods", ``Lismore Floods", \#NSWFloods, \#LismoreFloods, \#SydneyStorm, \#LismoreFloods2022, \#SydneyFloods, \#rainbomb, \#NSWFloods2022. There are 55,778 tweets from 18,773 unique users, including 1,437 geolocated tweets with longitude and latitude. 14,023 images and 2,625 videos shared by those tweets are stored separately for further potential analysis. This data captures how individuals and communities engaged with and responded to the disaster as it unfolded.

\textbf{Public submissions from 2022 NSW Flood Inquiry.} The Inquiry provided the researchers access to the submissions for research and analysis. Of the 1,494 submissions received, 1,250 are publicly available on the official website \cite{nsw_flood_inquiry_2022}. The researchers ensured that no identifying data was used in the analysis and respected the privacy of individuals who requested their submissions not be published. Most submissions were made shortly after the floods, with 80\% submitted in May and June. The comments focused on disaster experiences, recovery efforts, and suggestions to improve future preparedness, including emergency services, infrastructure, governance, and urban planning. Respondents included flood-affected residents, emergency personnel, local businesses, and community organizations, offering diverse perspectives. The survey also gathered multimedia evidence, such as 580 images from 252 submissions and location data from 148 postcodes, to better understand the disaster’s impact on communities. These insights were used to inform policy recommendations, enhance preparedness, and improve emergency response strategies.

\subsection{Topic Modelling Technique}
Our methodology uses topic modelling techniques separately on tweets and public submissions to extract distinct topics from each data source, allowing us to compare their thematic structures and identify differences or similarities in public concerns, needs, and situational dynamics for more informed decision-making. The most popular one, Latent Dirichlet Allocation (LDA), is a generative probabilistic model used for discovering latent topics in large text corpora by assuming that documents are mixtures of various topics, where each topic is characterised by a distribution over words \cite{blei2003latent}. 

We use Latent Dirichlet Allocation (LDA) to identify key topics in public submissions and social media data. Coherence scores \cite{roder2015exploring}, measured with $gensim$, help optimise the number of topics, ensuring clarity and relevance.
Data preprocessing is crucial for LDA topic modelling, as it reduces noise, standardises text, and ensures more meaningful topic extraction \cite{blei2003latent}. Our preprocessing pipeline consists of two main stages: text cleaning and tokenization.
First, text cleaning involves converting all characters to lowercase and removing irrelevant elements such as special characters, URLs, numbers, punctuation (except for periods), and Twitter-specific markers (e.g., @username). This cleaned text serves as input for the relevance index computation in Section \ref{index_measure}. 
Second, tokenization is performed at the sentence level to preserve context, followed by word-level tokenization while retaining meaningful N-grams (up to five words). Stopwords and excessively long words (over 17 letters) are removed to enhance topic coherence \cite{roder2015exploring}. After preprocessing, the dataset comprises 55,724 tweets and 1,450 public submissions, optimised for topic modelling.

\captionsetup[figure]{skip=0pt}
\begin{figure}[h]
     \centering
     \captionsetup[subfigure]{singlelinecheck=false,skip=0pt}
     \begin{subfigure}
         \centering
         \includegraphics[width=0.8\linewidth]{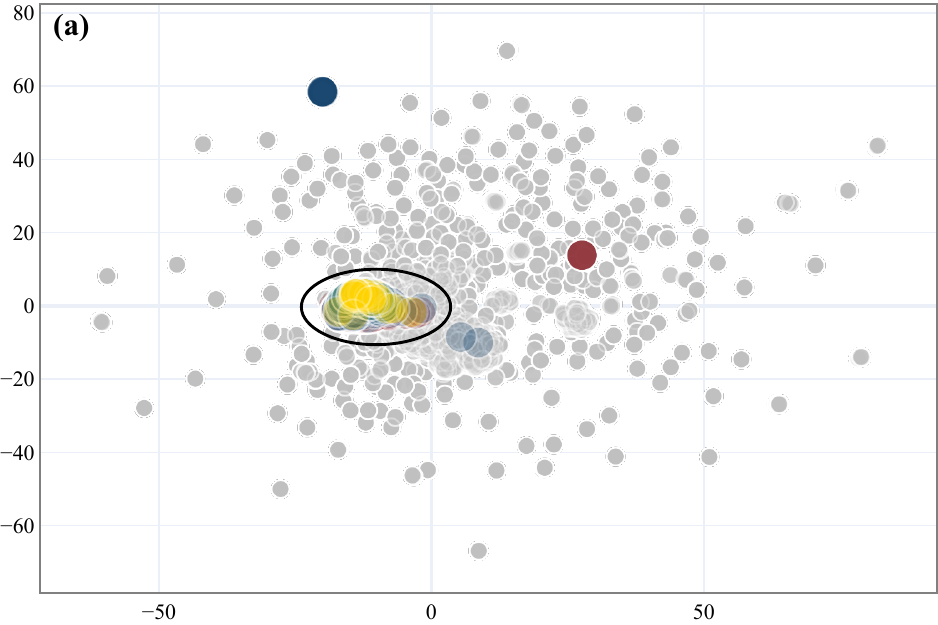}
     \end{subfigure}
     \begin{subfigure}
         \centering
         \includegraphics[width=0.8\linewidth]{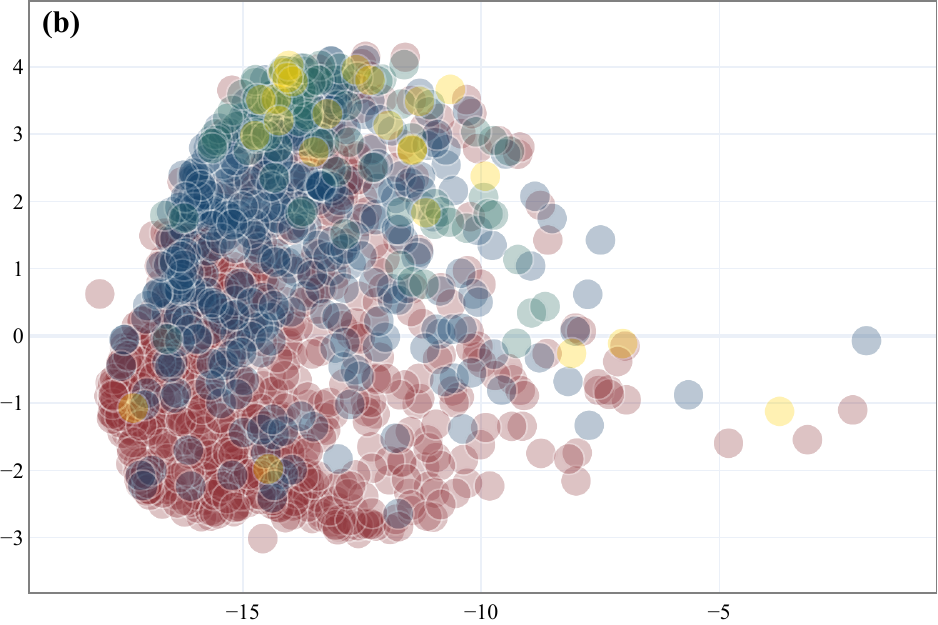}
     \end{subfigure}
     \includegraphics[width=0.8\linewidth]{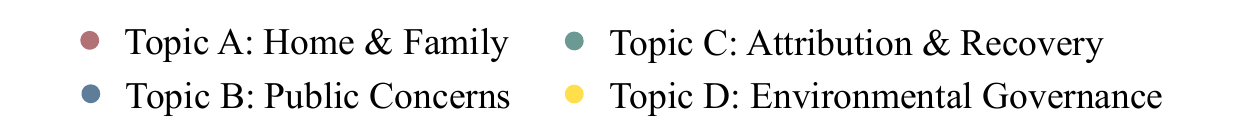}
     \vspace{5pt}
     \caption{\textbf{Visualisation of tweet and public submission embeddings using 2D UMAP}. (a) Grey points represent 55,724 tweets, while colored points indicate 1,450 public submissions. The black circle highlights the core cluster of submissions, which serve as a reference corpus for relevance filtering. (b) Topic distribution of public submissions within the black circle, categorized into Home \& Family, Public Concerns, Attribution \& Recovery, and Environmental Governance. For detailed topic descriptions see Table \ref{table:topic_modelling_tab}.} \label{fig:umap}
\end{figure}

\subsection{Relevance Index Design with LLMs} \label{index_measure}
In social media analysis of natural disasters, identifying relevant tweets helps to understand real-time situational awareness, emergency needs, and public sentiment \cite{vieweg2010microblogging,imran2015processing}. The openness of social media data introduces significant ``noise", with unrelated or low-relevance tweets. In contrast, public submissions and testimony at the inquiry are highly focused and directly relevant, representing ``all signal and no noise". To leverage this focused data, we used public submissions as a reference corpus to create a relevance index for tweets, utilising large language models (LLMs). 

As shown in Figure \ref{fig:umap}(a), public submissions (represented by coloured points) form natural, tight clusters in the 2D map, while tweets (shown in grey) display a more dispersed distribution. The black circle demarcates a region containing over 95\% of public submissions. Our manual analysis revealed that submissions falling outside this circle predominantly contained repetitive or insubstantial content. Given this observation, we focused on the submissions within the black circle (Figure \ref{fig:umap}(b)) as our reference corpus, as they demonstrated strong topical relevance to the flood events. To formalise our approach, we denote the embedding vector of a tweet (t) as \(\mathbf{v}_t \in \mathbb{R}^{1 \times d}\), where \(d\) represents the dimensionality of the embedding space. Similarly, we represent the reference corpus as a matrix \(\mathbf{R} \in \mathbb{R}^{n \times d}\), where \(n\) is the number of submissions.
\begin{equation} \label{equ2}
\mathbf{s} = \frac{\mathbf{v}_t \cdot \mathbf{R}^T}{\parallel \mathbf{v}_t \parallel \cdot \parallel \mathbf{R} \parallel_2}
\end{equation}
where \(\mathbf{s} \in \mathbb{R}^{1 \times n}\) is the vector of pairwise cosine similarities. \(\parallel \mathbf{v}_t \parallel\) is the Euclidean norm (L2 norm) of the tweet vector. \(\parallel \mathbf{R} \parallel_2\) is the L2 norms of the embedding matrix \(\mathbf{R}\). 

We treat the vector \(\mathbf{s}\) as the distribution of tweet similarity scores relative to the reference corpus and observe a negative skew. Since many statistical methods assume normality for consistent and interpretable outcomes, we apply the Box-Cox transformation to normalize the distribution, as shown in Equation \ref{equ3} \cite{box1964analysis,tabachnick2013using}. The transformed values are then inverted and scaled to ensure positivity and interpretability. The median of the adjusted distribution, \(\mathbf{y}_{scaled}\), serves as a robust relevance index for each tweet, with higher values indicating greater relevance to the flood disaster.

\begin{equation} \label{equ3}
\mathbf{y} = \begin{cases}
\frac{\mathbf{s}^{\lambda}-1}{\lambda}, \ \ \ \lambda \neq 0\\
\ln(\mathbf{s}), \ \ \  \lambda = 0
\end{cases} \ \Longrightarrow \ \ \ \ \mathbf{y}_{scaled} = -\frac{1}{\mathbf{y}}
\end{equation}

The Box-Cox transformation is effective for normalizing positive-valued data distributions, making it suitable for our relevance index. Although the Yeo-Johnson transformation \cite{yeo2000new} accommodates non-positive values, Box-Cox is preferred due to its well-established theoretical properties and prior success in disaster response studies \cite{entink2009box,xie2023changing}. Additionally, since our similarity scores are inherently positive, Box-Cox remains the optimal choice.

The LongT5 model \cite{guo2021longt5}  is a variant of the T5 transformer model we usd to generate embeddings for both tweets and public submissions, allowing for more effective similarity analysis. LongT5 is an extension of the original T5 model, designed to efficiently process longer text sequences while preserving contextual meaning.

Unlike standard transformer models, which struggle with lengthy inputs due to their quadratic memory complexity, LongT5 incorporates sparse attention mechanisms inspired by Longformer \cite{beltagy2020longformer}. These mechanisms enable the model to focus on both local word relationships and broader document-level context without excessive computational costs. 

This makes LongT5 particularly suitable for our study, where public submissions are typically long and structured, while tweets are short and fragmented. By applying LongT5 embeddings, we can effectively align social media discussions with structured public submissions, improving the relevance filtering of disaster-related content.

\section{Results}

\begin{figure}[h]
  \centering
  \includegraphics[width=\linewidth]{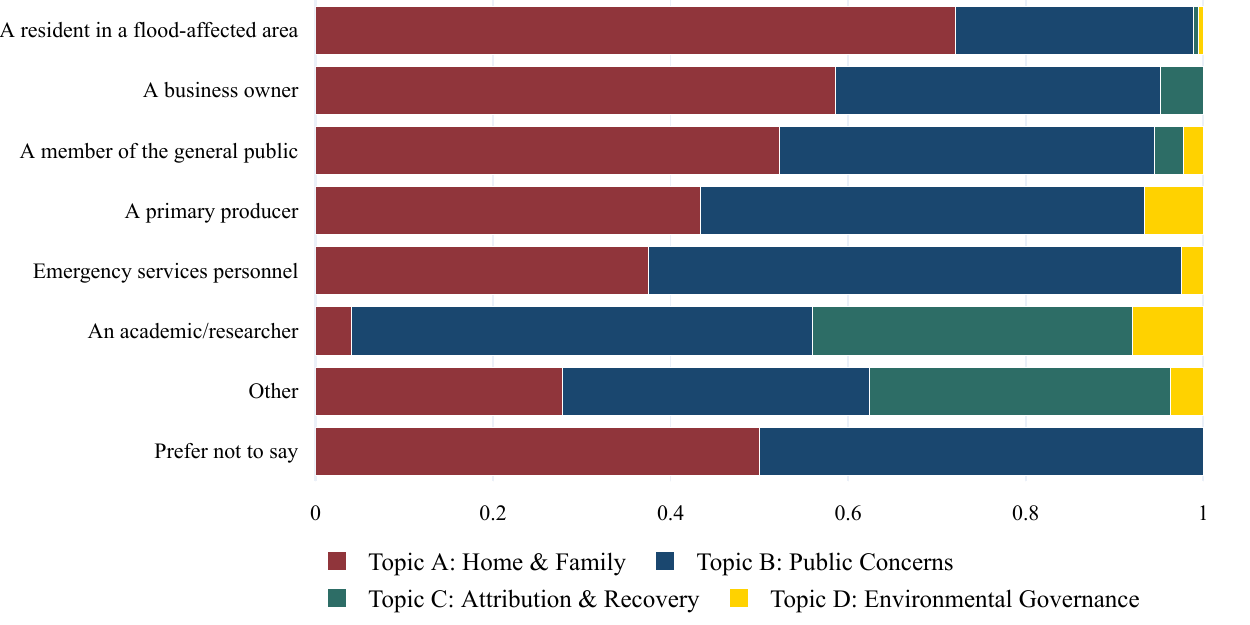}
  \vspace{0.1pt}
  \caption{\textbf{Submission topics by respondent type.} The figure illustrates the proportion of each topic—Home \& Family, Public Concerns, Attribution \& Recovery, and Environmental Governance—across different categories of submitters, including flood-affected residents, business owners, emergency personnel, and academics.}
  \label{fig:submitter_distribution}
\end{figure}

\subsection{Public Behaviour Patterns}
Latent Dirichlet Allocation (LDA) identifies four topics in 1,450 public submissions and six in 55,724 tweets. Table \ref{table:topic_modelling_tab} summarises each topic’s title, distribution, and keywords. Since LDA primarily generates unigram keywords, limiting interpretability, we apply Revealed Comparative Advantage (RCA) to enhance topic representation. First, we extract overlapping N-grams within each dataset (public submissions and tweets). Then, treating topics as ``countries" and N-grams as ``products", we calculate RCA scores to determine each N-gram’s relative importance.

The RCA for a specific N-gram \(i\) in a topic \(t\) is calculated as follows:
\begin{equation} \label{equ4}
RCA_{ti} = \frac{\frac{f_{ti}}{\sum_{k \in K}f_{tk}}}{\frac{\sum_{t \in T}f_{ti}}{\sum_{t \in T}\sum{k \in K}f_{tk}}}
\end{equation}
where \(f_{ti}\) is the frequency of an N-gram \(i\) in topic \(t\), and \(f_{tk}\) is the frequency of all N-grams \(K\) in topic \(t\). The numerator captures the share of an N-gram \(i\) within topic \(t\), while the denominator measures the overall share of an N-gram \(i\) across all topics.
The final step is to rank the N-grams for each topic in descending order, from 5-grams to unigrams, prioritising longer phrases as they tend to encapsulate substantial information. We manually select the ten representative phrases for each topic, ensuring the interpretability and semantic coherence of the topics.

\setlength{\heavyrulewidth}{1pt} 
\setlength{\lightrulewidth}{1pt} 
\setlength{\cmidrulewidth}{0.3pt}  
\newcolumntype{M}[1]{>{\centering\arraybackslash}m{#1}}

\begin{table*}[h]
\centering
  \caption{\textbf{Summary of topic modelling results for public submissions and social media tweets.} The table presents the main topics, their frequency (number and percentage of total), and representative keywords, highlighting differences in discourse between structured submissions and real-time social media discussions.}
  \label{table:topic_modelling_tab}
  \begin{tabular}{cM{1cm}M{2cm}M{1cm}M{9cm}}
    \toprule
    &\textbf{Topics}&\textbf{Title}&\textbf{\#/\%}&\textbf{10 Keywords}\\
    \midrule
    \multirow{4}{*}{\rotatebox[origin=c]{90}{\parbox[c]{4cm}{\centering \textbf{Public Submissions}}}} & \multicolumn{1}{c}{\makecell{\includegraphics[width=0.035\linewidth]{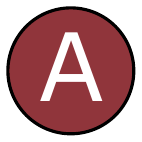}}} & \multicolumn{1}{c}{Home \& Family}  & \multicolumn{1}{c}{891/61\%} & \multicolumn{1}{c}{\parbox{9cm}{evacuation centre, insurance companies, lack of support, phone or internet, people with disabilities, people and animals, food and clothing, power and water, women and children, voluntary house raising scheme}} \\
    \cmidrule(lr){2-5} 
    & \multicolumn{1}{c}{\makecell{\includegraphics[width=0.035\linewidth]{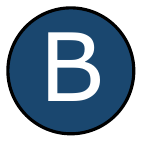}}} & \multicolumn{1}{c}{Public Concerns}  & \multicolumn{1}{c}{448/31\%} & \multicolumn{1}{c}{\parbox{9cm}{ses management, response plan, defence forces, public servants, climate scientists, local citizens, require assistance, beef cattle, search and rescue, loss of income}} \\
    \cmidrule(lr){2-5} 
    & \multicolumn{1}{c}{\makecell{\includegraphics[width=0.035\linewidth]{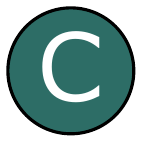}}} & \multicolumn{1}{c}{Attribution \& Recovery}  & \multicolumn{1}{c}{87/6\%} & \multicolumn{1}{c}{\parbox{9cm}{architecture, emergency management committee, vulnerable community members, response to and recovery, programs applying to proposed future, current laws emergency management plans, causes of and factors contributing, weather climate change and human, land use planning and management, adapt to future flood risks}} \\
    \cmidrule(lr){2-5} 
    & \multicolumn{1}{c}{\makecell{\includegraphics[width=0.035\linewidth]{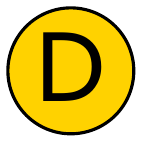}}} & \multicolumn{1}{c}{Environmental Governance}  & \multicolumn{1}{c}{24/2\%} & \multicolumn{1}{c}{\parbox{9cm}{advises, react, resource management, wildlife corridors, rainfall patterns, water systems, native species, land values, plants and animals, address climate change}} \\

    \midrule
    \multirow{6}{*}{\rotatebox[origin=c]{90}{\parbox[c]{4.5cm}{\centering \textbf{Social Media Tweets}}}} & \multicolumn{1}{c}{\makecell{\includegraphics[width=0.035\linewidth]{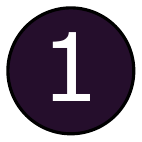}}} & \multicolumn{1}{c}{Election \& Politics}  & \multicolumn{1}{c}{10,871/20\%} & \multicolumn{1}{c}{\parbox{9cm}{lnp (liberal national party), action, vote, blame, abandoned, next election, still homeless, city council, ses police, people died floods}} \\
    \cmidrule(lr){2-5} 
    & \multicolumn{1}{c}{\makecell{\includegraphics[width=0.035\linewidth]{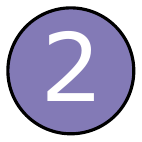}}} & \multicolumn{1}{c}{Rescue \& Donate}  & \multicolumn{1}{c}{10,765/19\%} & \multicolumn{1}{c}{\parbox{9cm}{help people, lives lost, insurance claims, emergency services, please help, save lives, flood relief, recovery efforts, please donate, lost everything floods}} \\
    \cmidrule(lr){2-5} 
    & \multicolumn{1}{c}{\makecell{\includegraphics[width=0.035\linewidth]{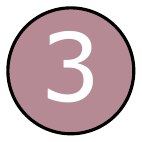}}} & \multicolumn{1}{c}{Weather Information}  & \multicolumn{1}{c}{9,362/17\% } & \multicolumn{1}{c}{\parbox{9cm}{water levels, storm water, bad news, nsw disaster, heavy rainfall, weather forecast, bureau meteorology, floods rain, flash flooding, high tide}} \\
    \cmidrule(lr){2-5} 
    & \multicolumn{1}{c}{\makecell{\includegraphics[width=0.035\linewidth]{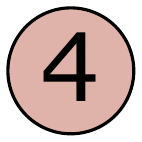}}} & \multicolumn{1}{c}{Public Sentiment}  & \multicolumn{1}{c}{9,114/16\%} & \multicolumn{1}{c}{\parbox{9cm}{adf (Australian Defence Force), taxpayers, sandbags, deputy premier, bridget mckenzie, emergency response, people dying, houses destroyed, record breaking floods, left homeless floods}} \\
    \cmidrule(lr){2-5} 
    & \multicolumn{1}{c}{\makecell{\includegraphics[width=0.035\linewidth]{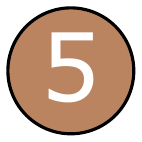}}} & \multicolumn{1}{c}{Advertising}  & \multicolumn{1}{c}{8,078/14\%} & \multicolumn{1}{c}{\parbox{9cm}{taylor, michael, acting, players, affair, rock, song, newspaper, radio, playing}} \\
    \cmidrule(lr){2-5} 
    & \multicolumn{1}{c}{\makecell{\includegraphics[width=0.035\linewidth]{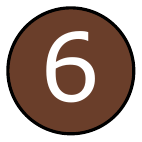}}} & \multicolumn{1}{c}{News Live Updates}  & \multicolumn{1}{c}{7,534/14\%} & \multicolumn{1}{c}{\parbox{9cm}{cabinet, combat, invasion ukraine, covid deaths, national disaster, scott morrison, news live, press conference, disaster payments, australia news live updates}} \\
  \bottomrule
\end{tabular}
\end{table*}

\begin{figure}[h]
  \centering
  \includegraphics[width=\linewidth]{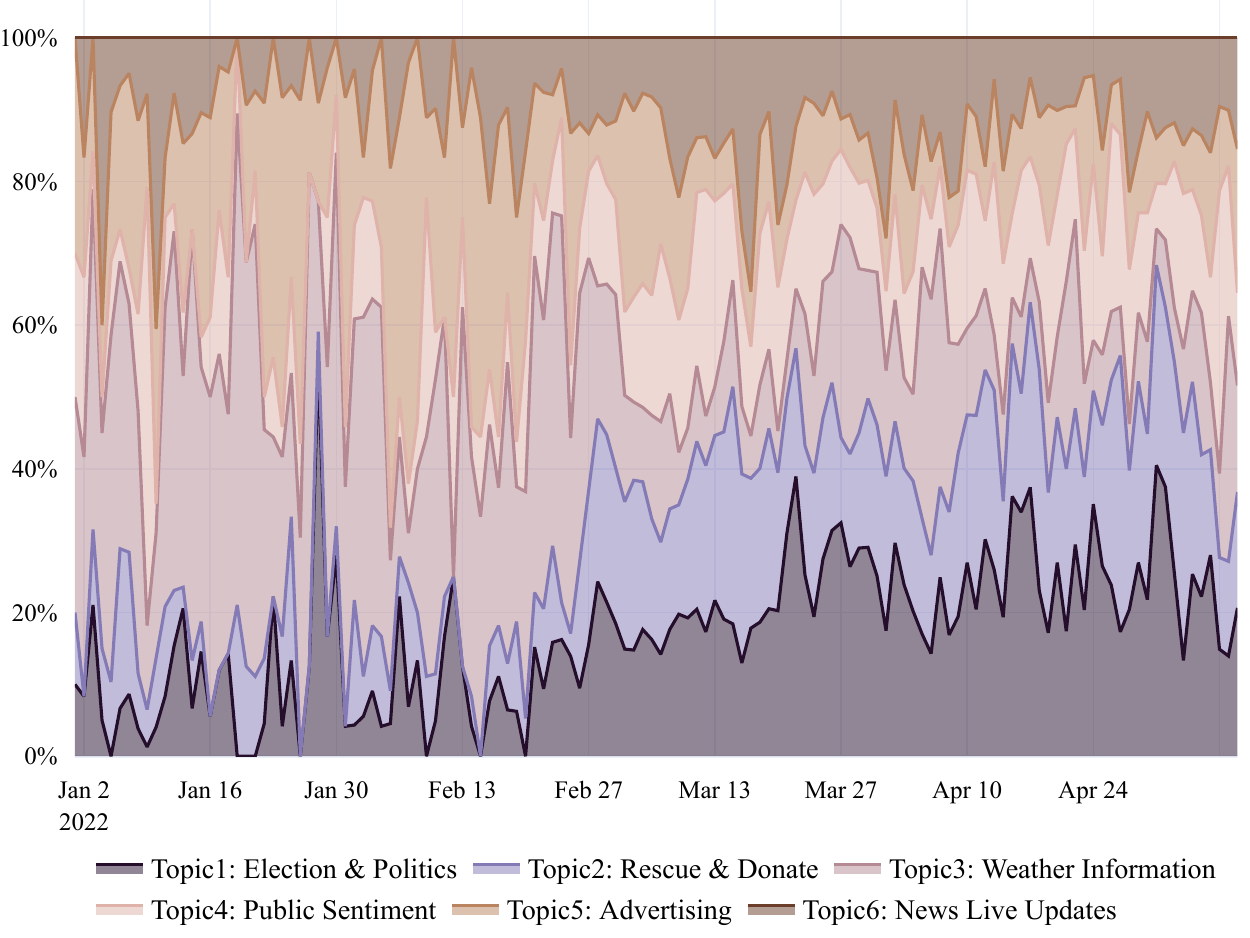}
  \vspace{0.1pt}
  \caption{\textbf{Topic distribution of daily tweet volume.} Initially, weather updates and rescue efforts dominated, peaking during major flood events. Over time, discussions shifted toward long-term recovery, political debates on future preparedness, and increasing public sentiment, while real-time news coverage declined.}
  \label{fig:tweet_topic_distribution}
\end{figure}

\begin{figure}[h]
  \centering
  \includegraphics[width=0.9\linewidth]{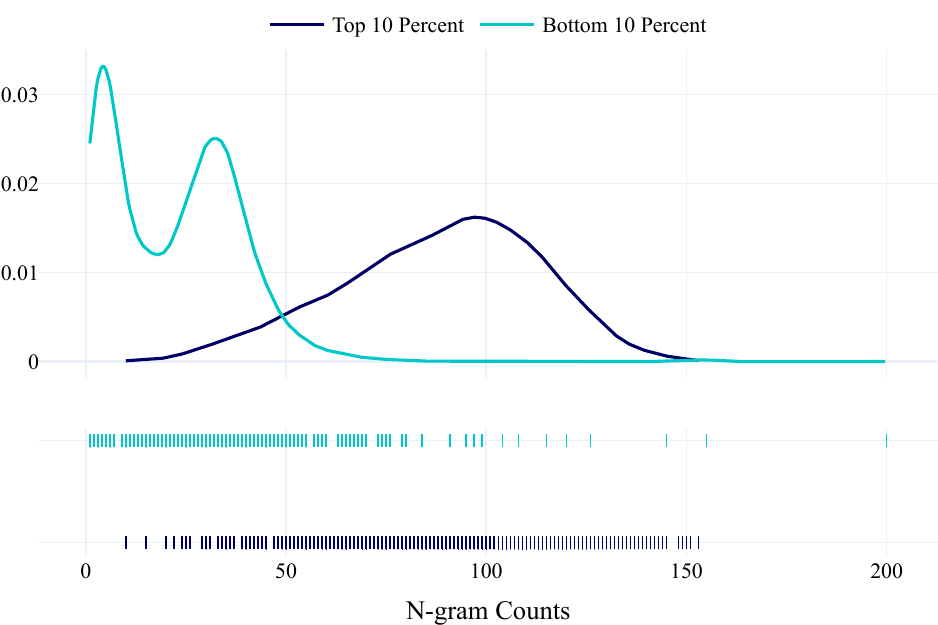}
  \vspace{1pt}
  \caption{\textbf{Density distribution of N-gram counts in the top 10\% and bottom 10\% of tweets ranked by the relevance index. }The distribution shows that highly relevant tweets (top 10\%) tend to have higher N-gram counts, while less relevant tweets (bottom 10\%) exhibit a more varied and lower-frequency distribution.}
  \label{fig:ngram_distribution}
\end{figure}

Most public submissions are categorised under ``Home \& Family", which focuses on personal experiences of the disaster. Figure \ref{fig:submitter_distribution} shows that most submissions are from individuals in directly impacted regions, particularly the Northern Rivers area, where the flooding had the most severe impact (see Figure \ref{fig:public_geo}). This topic primarily focuses on issues related to personal safety and basic needs, including the utilisation of evacuation centres, challenges in accessing insurance, deficiencies in support services, and the provision of essentials such as food, clothing, power, and water. It also highlights specific concerns faced by vulnerable groups, such as people with disabilities, women, and children, as well as the welfare of both people and animals and community-based initiatives like the voluntary house-raising scheme to mitigate future flood risks. 

Another big Topic B, ``Public Concerns", reflects broader concerns related to SES management, response planning, involvement of defence forces, and roles of public servants and climate scientists. It includes more submissions from the emergency services personnel, highlighting their pivotal role in shaping discussions on disaster response. In Topic C, ``Attribution \& Recovery", most submissions originated from academia with attachments, a trend also evident in the geographical distribution, shifting from the Northern Rivers region towards Sydney. It pays more attention to themes related to both the structural and administrative aspects of disaster response to enhance resilience and adapt to future flood risks. The ``Environmental Governance" topic D emphasises issues related to managing natural resources, highlighting the importance of wildlife corridors, water systems, and native species conservation. This topic adopts a long-term perspective, emphasising the interconnection between the environment and human communities, and explores strategies to mitigate the occurrence of disasters fundamentally.

Social media tweets illustrate distinctive patterns from public submissions, with the most prevalent being ``Election \& Politics", followed by ``Rescue \& Donate",  ``Weather Information", and ``Public Sentiment". These results suggest a broader spectrum of engagement on social media, including political discussions, calls for assistance, and informational sharing. Notably, public submissions are more focused on immediate evacuation and support, whereas social media conversations encompass various topics, including political discourse, public sentiment, and live news updates. This disparity highlights the difference in the nature of concerns shared through formal submissions versus those expressed on social media, reflecting different levels of engagement and topics of interest among individuals during flood events. 

The distribution of tweets across social media is relatively balanced in volume. Content-wise, the first four topics are closely related to floods, whereas the last two topics exhibit more ``noise" with unrelated keywords. In Figure \ref{fig:tweet_topic_distribution}, except for Weather Information Topic 3 experienced a decline in proportion during the flooding period, the other three flood-related topics (Topics 1, 2, and 4) initially represented a smaller proportion but demonstrated a significant increase during the flooding period. Topic 5 exhibited the opposite trend. Geographically (see Figure \ref{fig:tweets_geo}), the first three topics predominantly originate from areas directly affected by the disaster, whereas the last two topics include more tweets from locations distant from the disaster-affected regions. Notably, the fourth topic contains significant criticism regarding the inadequate rescue efforts of the Australian Defence Force despite being geographically linked to Victoria, an area not impacted by the floods. 

\begin{figure*}
  \includegraphics[width=\textwidth]{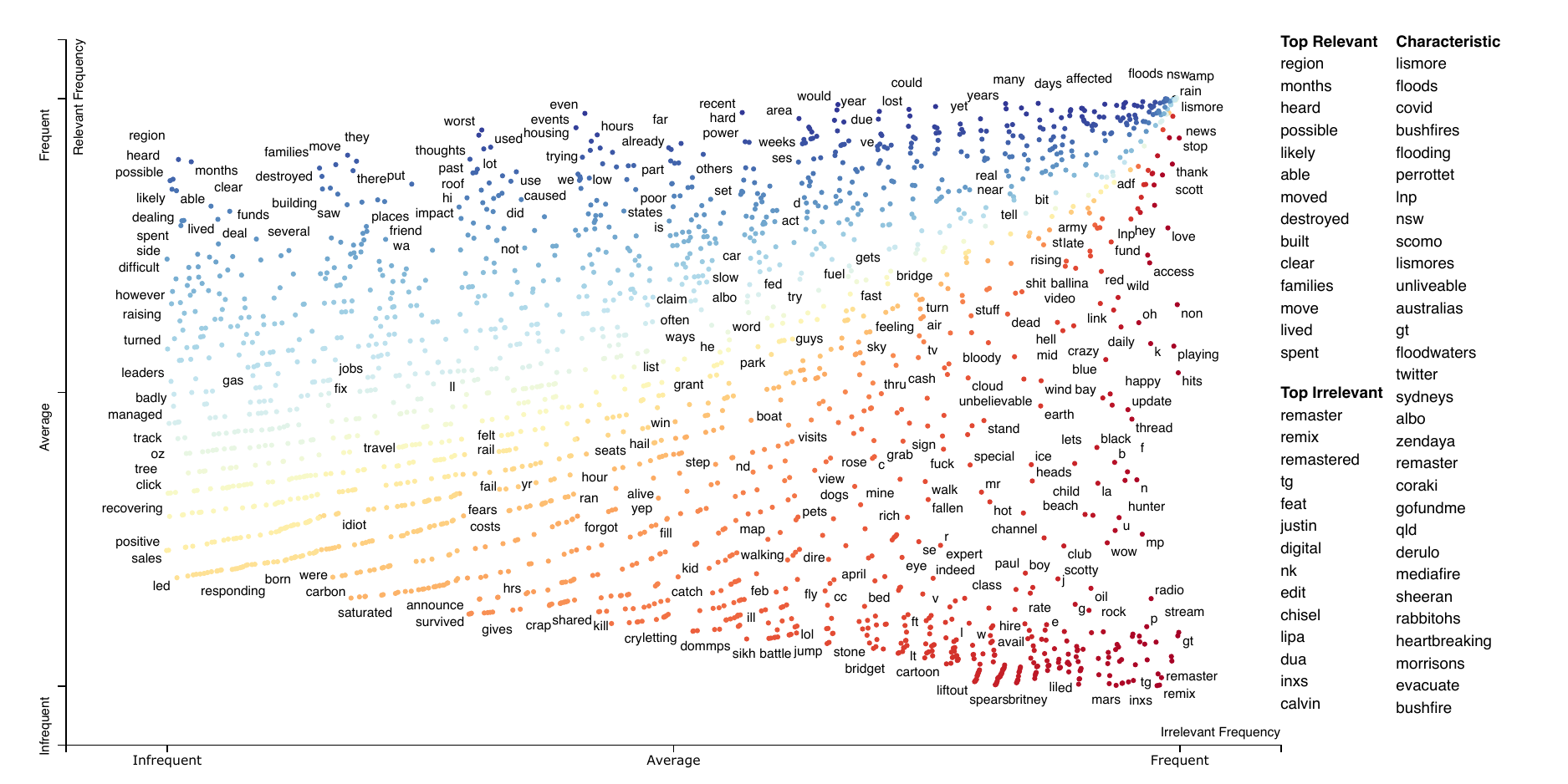}
  \vspace{0.5pt}
  \caption{\textbf{Linguistic Patterns of Relevant and Irrelevant Social Media Posts.} 
  This scatterplot displays the distribution of terms in flood-related posts, with top relevant terms (blue) indicating words most strongly associated with flood discussions 
  and top irrelevant terms (red) representing words likely to be noise or unrelated content. 
  The most characteristic terms (right) are those that appear most frequently across the dataset, 
  capturing key topics and themes present in the corpus.}
  \label{fig:tweet_scattertext}
\end{figure*}

\subsection{Performance of Relevance Index}
We evaluate the effectiveness of the relevance index by analyzing the top and bottom 10\% of ranked tweets, focusing on two key aspects: content informativeness and keyword frequency distribution using the $scattertext$ tool.

To measure informativeness, we examine the N-gram counts of each tweet, where higher counts indicate greater informativeness. Figure \ref{fig:ngram_distribution} compares the density distributions of N-gram counts for high-relevance (top 10\%) and low-relevance (bottom 10\%) tweets. The high-relevance group exhibits a concentrated distribution of N-grams, suggesting a more focused use of key terms. In contrast, the low-relevance group shows a flatter distribution at higher N-gram counts, implying more dispersed and infrequent use of relevant terms. This contrast highlights that tweets with higher relevance scores tend to use key phrases more consistently.
We analyse linguistic differences between relevant and irrelevant tweets on flood-related content using keyword frequency and association metrics. Figure \ref{fig:tweet_scattertext}, generated using Scattertext \cite{kessler2017scattertext}, highlights distinct patterns. Relevant tweets frequently include terms such as ``region", ``families", ``months", ``clear", and ``destroyed", reflecting location-specific impacts, personal narratives, and temporal references. In contrast, irrelevant tweets are dominated by unrelated music and entertainment references such as ``remaster", ``remix", ``lipa", and ``inxs", indicating off-topic discussions. Additionally, keywords like ``Lismore", ``floods", ``bushfires", ``covid", and ``Perrottet" suggest that relevant tweets often incorporate broader contextual discussions, including government response and comparisons to other natural disasters. This analysis reinforces the effectiveness of the relevance index in distinguishing flood-related content from extraneous discussions. The relevance index effectively filters irrelevant content, as demonstrated by the distinct linguistic patterns and the identification of off-topic terms within the dataset.




\section{Conclusion}
This research builds on analytical work commissioned by the NSW Government following the 2022 floods, extending those insights to enhance disaster response and resilience planning. Our findings align with the recommendations of the independent NSW Flood Inquiry and the government's subsequent response, particularly in strengthening flood preparedness, emergency coordination, and public communication \cite{govt_response_2022}. By integrating social media and public submissions, this study offers new analytical tools that support timely decision-making for emergency services and policymakers, aligning with the inquiry’s recommendations on risk management and inter-agency coordination.

Our key contributions include:
\begin{itemize}
    \item \textbf{AI-driven filtering:} The Relevance Index, leveraging LLM embeddings, improves crisis communication by prioritizing disaster-related social media content and reducing noise.
    \item \textbf{Topic-based insights:} LDA topic modeling reveals differences between structured public submissions and the fragmented yet dynamic discourse on social media.
    \item \textbf{Practical applications:} Findings support real-time crisis management and long-term policy planning, aiding emergency responders and government agencies.
\end{itemize}

Future research should expand the dataset scope, incorporating additional public submissions and applying multimodal analysis, such as combining textual and visual content, for deeper behavioural insights. Advancements in these areas could further refine AI-driven disaster analysis, improving situational awareness, emergency response, and resilience planning.

\appendix




\section*{Appendix}
\renewcommand{\figurename}{Figure A.}
\setcounter{figure}{0}

\begin{figure}[H]
\centering
\renewcommand*{\arraystretch}{0}
\setlength{\tabcolsep}{0pt} 
\begin{tabular}{cc}
\includegraphics[scale=0.25]{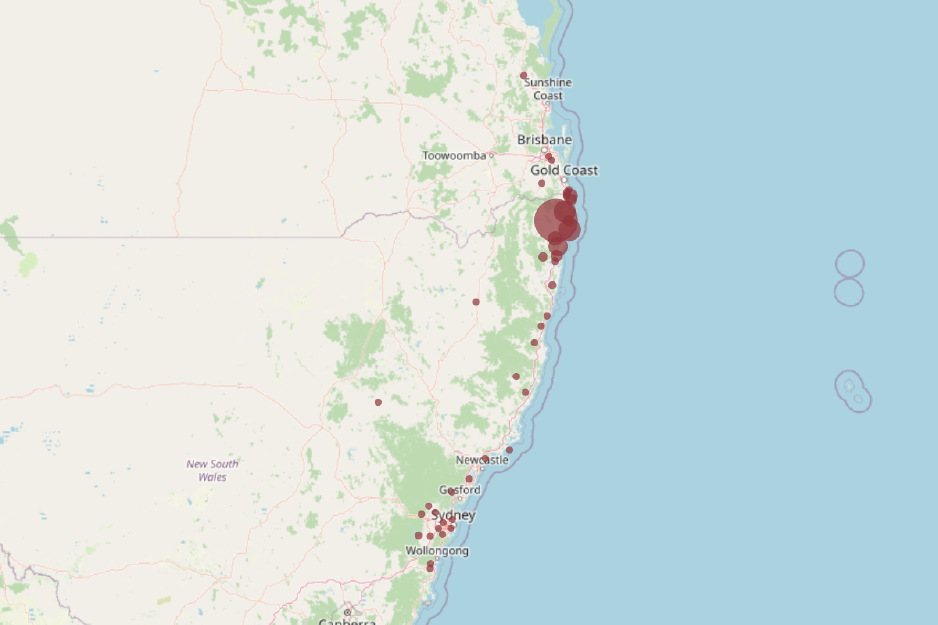}    & 
\includegraphics[scale=0.25]{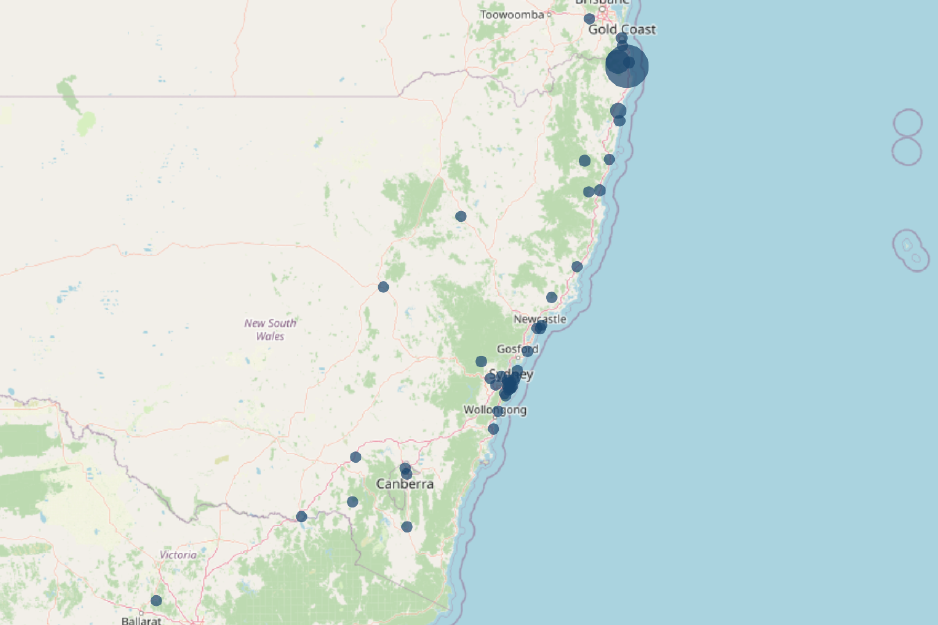}  \\
\includegraphics[scale=0.25]{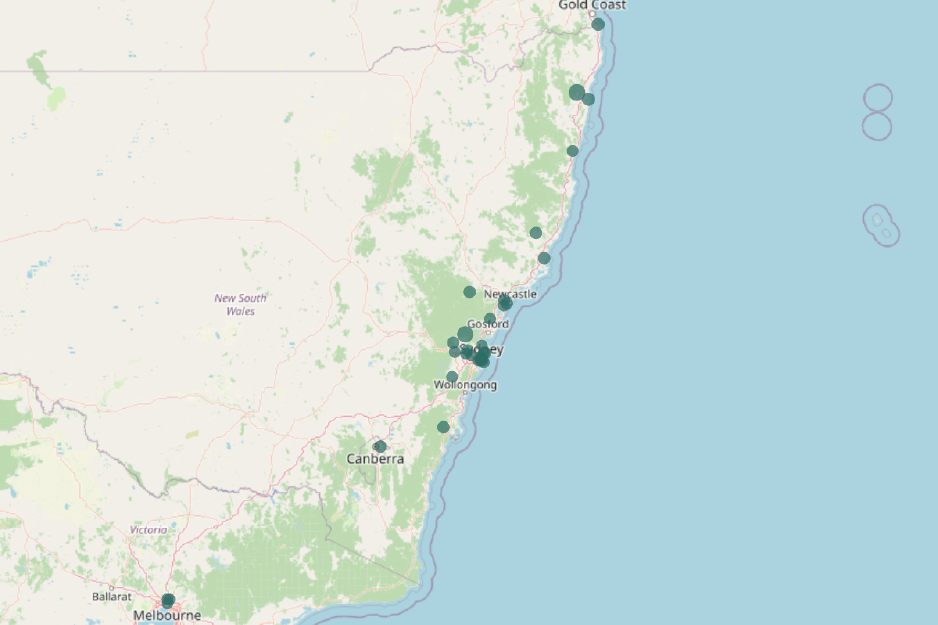}   & 
\includegraphics[scale=0.25]{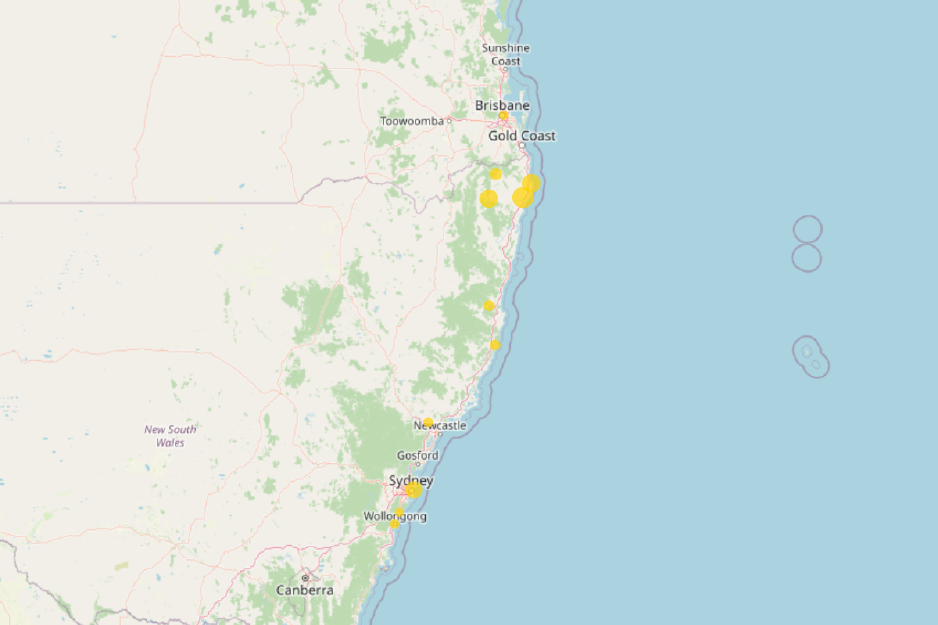}
\end{tabular}
\vspace{0pt}
\includegraphics[width=\linewidth]{Figures/geo_submissions_legend.pdf}
\vspace{5pt}
\caption{Geographical distribution of public submission by topics. \textmd{Find detailed  topic description in Table \ref{table:topic_modelling_tab}}}
\label{fig:public_geo} 
\end{figure}

\begin{figure}[H] 
\centering
\renewcommand*{\arraystretch}{0} 
\setlength{\tabcolsep}{0pt} 
\begin{tabular}{ccc}
\includegraphics[scale=0.17]{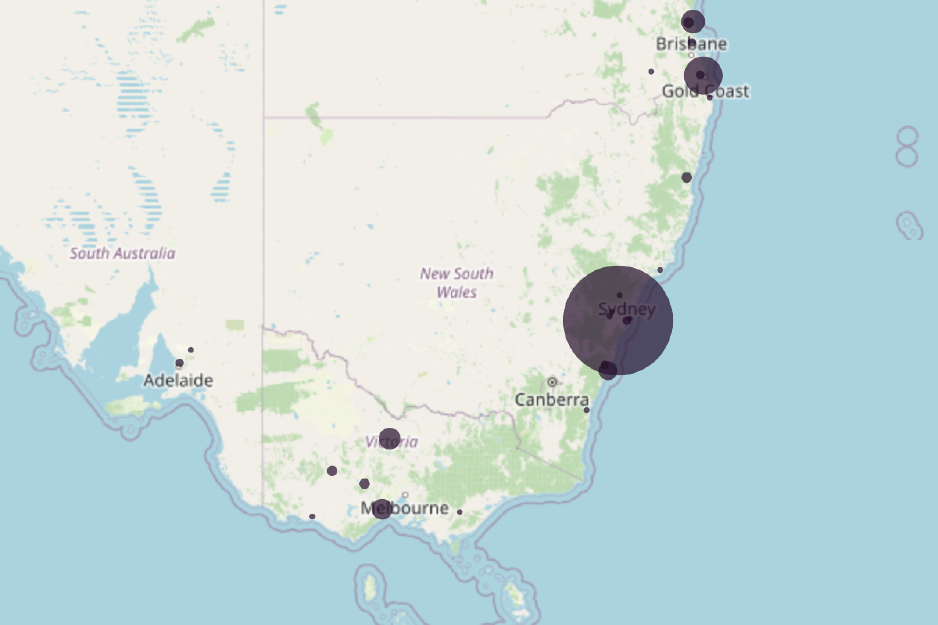} & 
\includegraphics[scale=0.17]{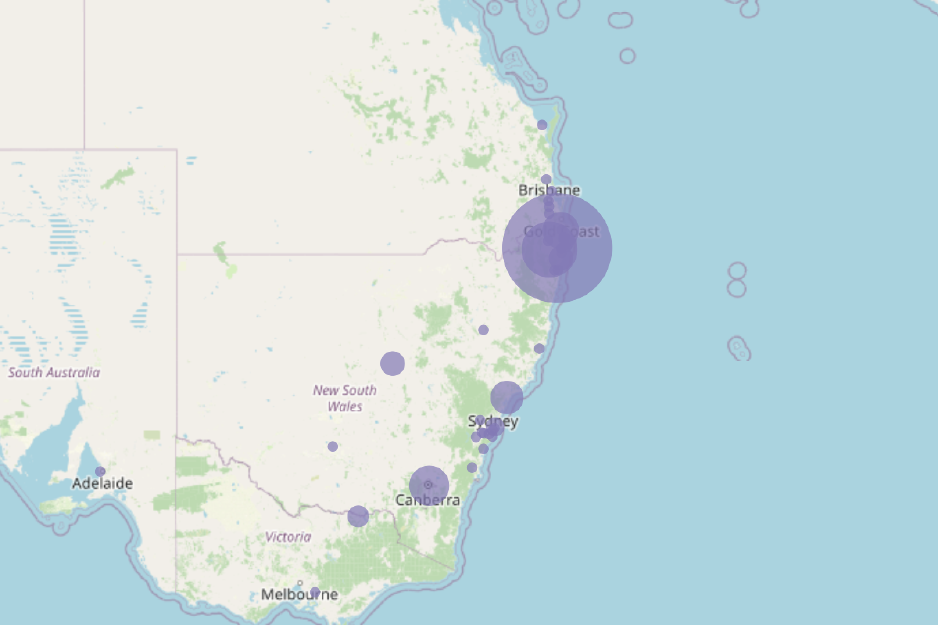} & 
\includegraphics[scale=0.17]{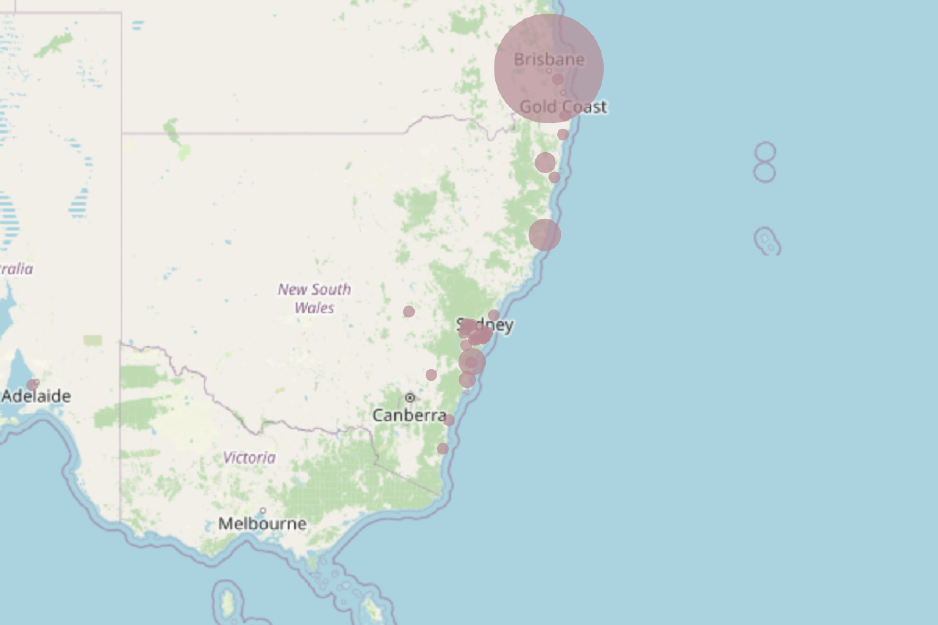} \\
\includegraphics[scale=0.17]{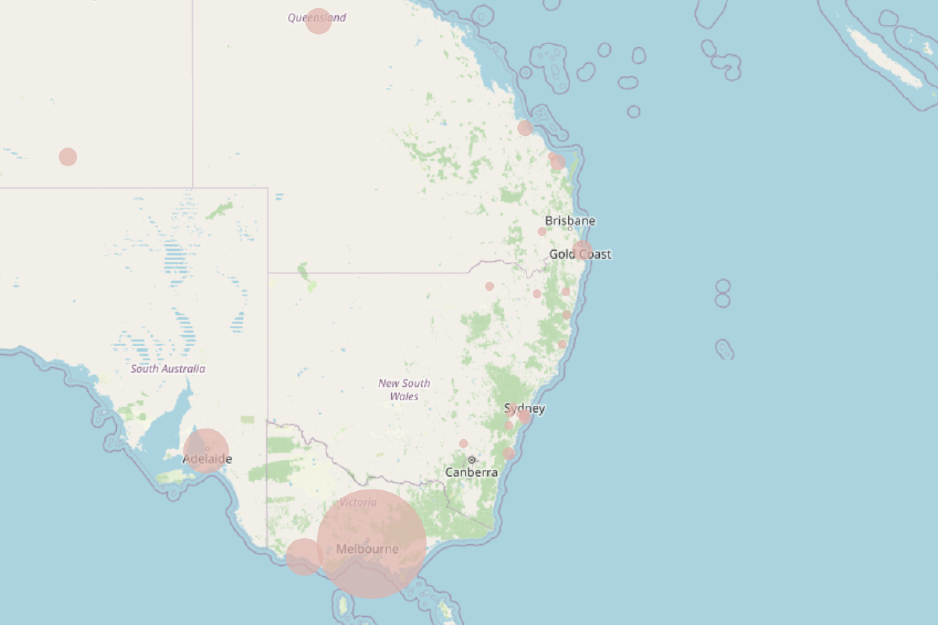} & 
\includegraphics[scale=0.17]{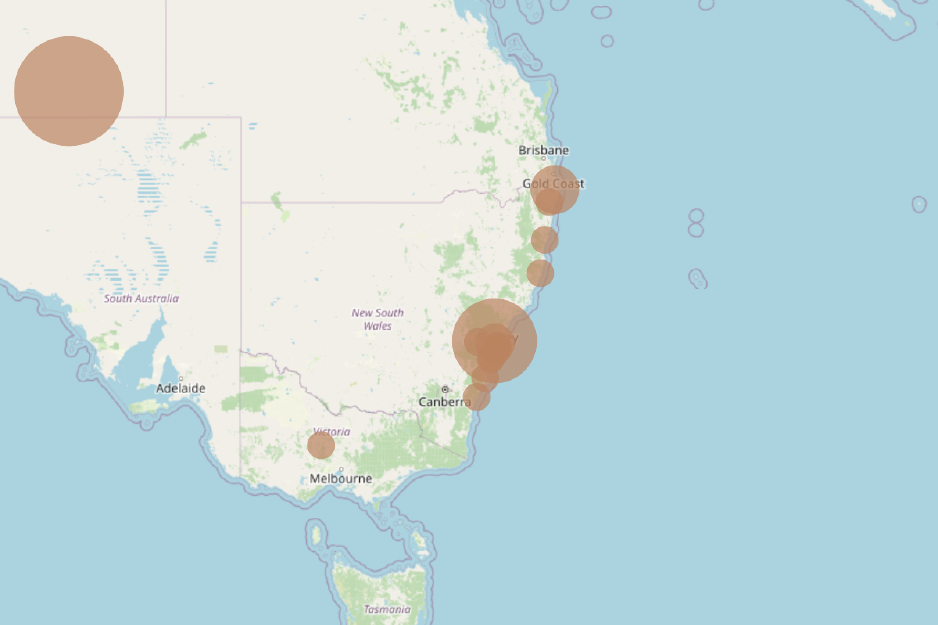} & 
\includegraphics[scale=0.17]{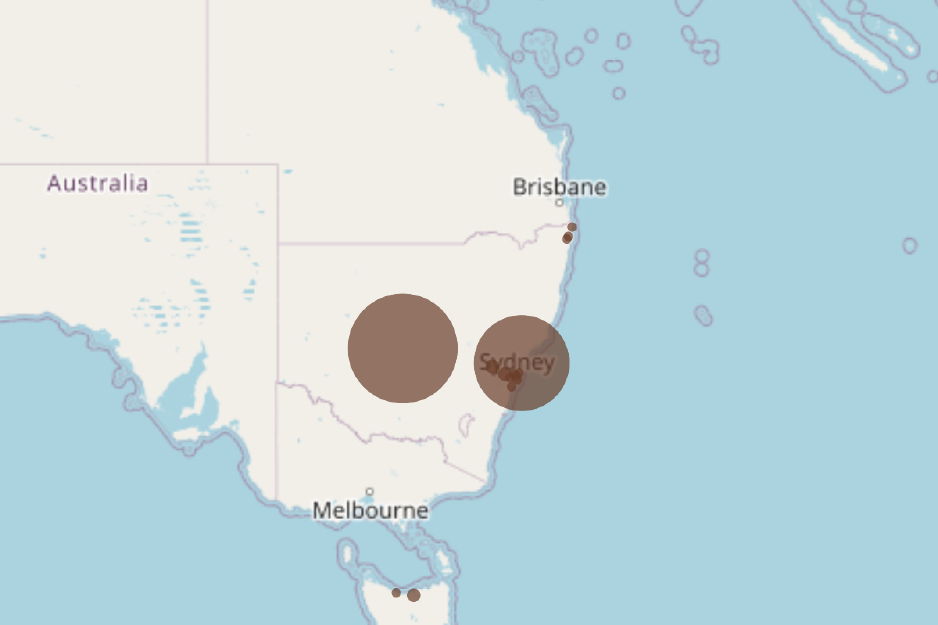} \\
\end{tabular}
\vspace{0pt}
\includegraphics[width=\linewidth]{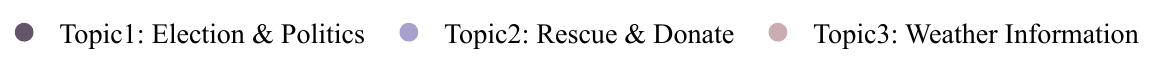} 
\includegraphics[width=\linewidth]{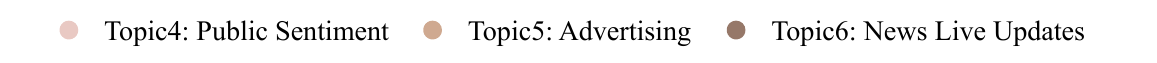} 
\vspace{1pt}
\caption{Geographical distribution of social media tweets by topics. \textmd{Find detailed topic description in Table \ref{table:topic_modelling_tab}}}
\label{fig:tweets_geo}
\end{figure}

\bibliographystyle{named}
\bibliography{ijcai25}

\end{document}